\documentclass{l4dc2024}
\usepackage{wrapfig}
\usepackage{lipsum}
\usepackage{amsmath}
\usepackage{enumitem}
\usepackage{cleveref}
\usepackage{float}
\newcommand{\PP}{\mathbb{P}}
\newcommand{\algo}{\textsc{MCMC-BO}}

\title[high dimensional Bayesian optimization with MCMC]{Improving sample efficiency of high dimensional Bayesian optimization with MCMC}
\usepackage{times}

\author{%
 \Name{Zeji Yi}\normalfont \textsuperscript{1}\textsuperscript{*} \Email{zejiy@andrew.cmu.edu}\\
 \Name{Yunyue Wei}\normalfont\textsuperscript{1}\textsuperscript{*} \Email{weiyy20@mails.tsinghua.edu.cn}\\
 \Name{Chu Xin Cheng}\normalfont\textsuperscript{2}\textsuperscript{*} \Email{ccheng2@caltech.edu}\\
 \Name{Kaibo He}\normalfont\textsuperscript{1} \Email{hkb21@mails.tsinghua.edu.cn}\\
 \Name{Yanan Sui}\normalfont\textsuperscript{1} \Email{ysui@tsinghua.edu.cn}\\
 \addr \textsuperscript{1}Tsinghua University, \textsuperscript{2}California Institute of Technology.  \\
 \footnotetext{*These authors contributed equally to this work. Code and Appendices are available at \url{https://drive.google.com/drive/folders/1fLUHIduB3-pR78Y1YOhhNtsDegaOqLNU?usp=sharing}.}
}

\begin{document}

\maketitle

\begin{abstract}
Sequential optimization methods are often confronted with the curse of dimensionality in high-dimensional spaces. Current approaches under the Gaussian process framework are still burdened by the computational complexity of tracking Gaussian process posteriors and need to partition the optimization problem into small regions to ensure exploration or assume an underlying low-dimensional structure. With the idea of transiting the candidate points towards more promising positions, we propose a new method based on Markov Chain Monte Carlo to efficiently sample from an approximated posterior. We provide theoretical guarantees of its convergence in the Gaussian process Thompson sampling setting. We also show experimentally that both the Metropolis-Hastings and the Langevin Dynamics version of our algorithm outperform state-of-the-art methods in high-dimensional sequential optimization and reinforcement learning benchmarks.
\end{abstract}


\section{Introduction}
With broad applications in real-world engineering problems, black-box function optimization is an essential task in machine learning \citep{snoek2012practical, hernandez2017parallel}. These non-convex sequential optimization problems often lack gradient information. Bayesian optimization (BO) is a popular sampling-based online optimization approach for solving expensive optimization problems. It has been successfully applied to problems such as online learning and sequential decision-making. BO builds a surrogate model for modeling the objective function and optimizes the acquisition function to propose new samples.

Similar to many other numerical problems, BO algorithms are also susceptible to the curse of dimensionality. The search space would grow exponentially large as the function dimension increases and become intractable under a limited computation budget. Common acquisition functions also tend to over-explore the uncertainty boundary region and lack exploitation in high dimensional input space \citep{oh2018bock}. Recent developments in high dimensional BO include constructing trust regions and space partitions to improve the probability of sampling in promising regions, which effectively ameliorates the problem of over-exploration \citep{eriksson2019scalable, wang2020learning}. To evaluate the acquisition function on a continuous domain, these methods often discretize the search space using Sobolev sequence \citep{sobol1967distribution}. However, this kind of discretization could be inadequate in high dimensional space as the size of the discretization set is limited, hindering the exploitation of potential good regions.

\begin{wrapfigure}{r}{0.5\textwidth}
  \begin{center}
\includegraphics[width=0.48\textwidth]{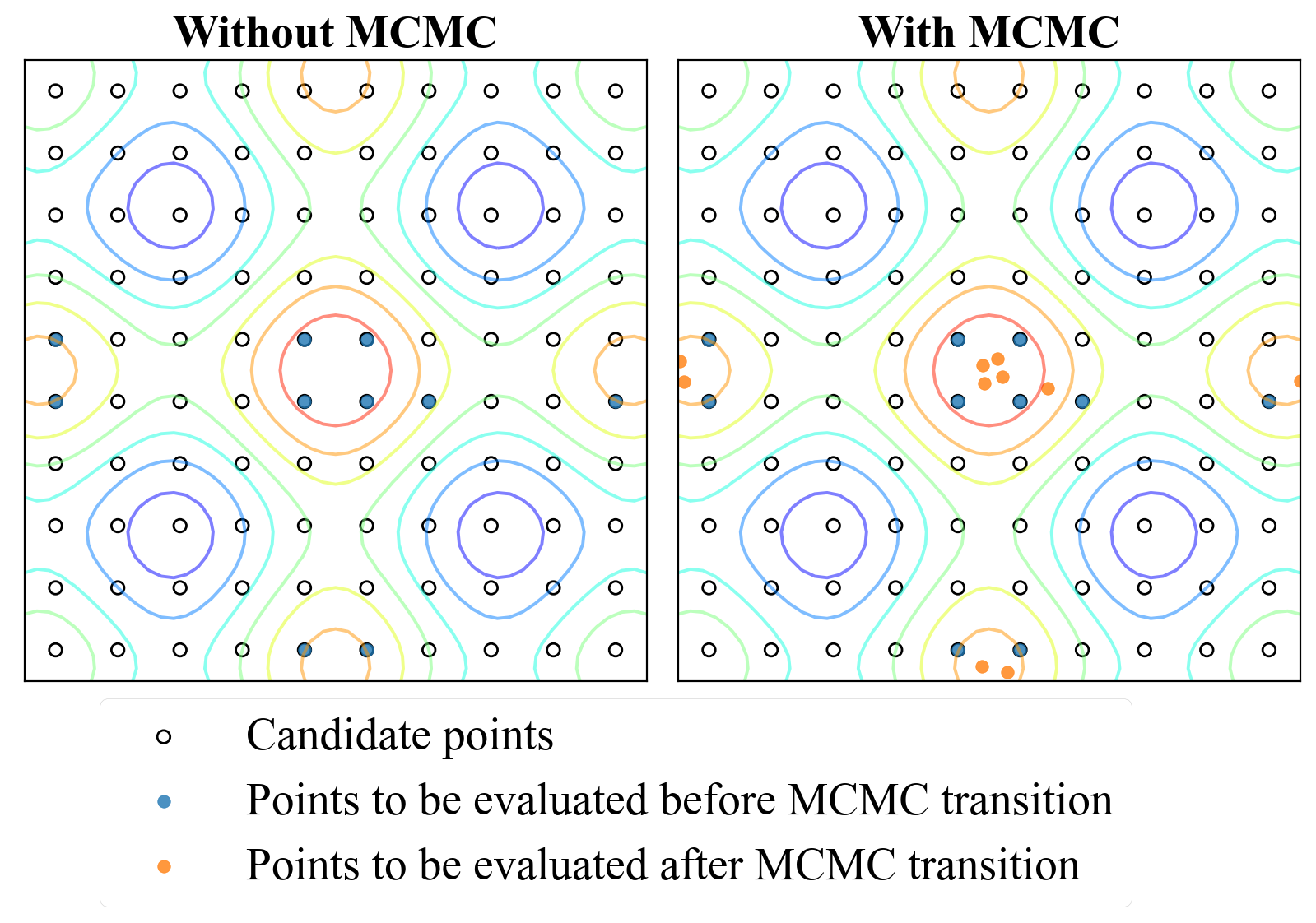}
  \end{center}
  \vspace{-20pt}
\caption{Illustration of \algo. The contours are 2d Rastrigin function. (\textbf{Left}): BO algorithms propose points to be sampled. The optimization performance is restricted by insufficient discretization. (\textbf{Right}): Points are adjusted by \algo~, reaching regions with higher value.}
  \label{fig:algo}
  \vspace{-20pt}
\end{wrapfigure}
In our setting, we investigate probabilistic acquisition such as Thompson sampling (TS) for optimizing unknown function modeled with Gaussian processes (GP). Yet, directly applying algorithms that perform well in low-dimensional problems to high-dimensional domains often suffers due to the high-computation demand and the phenomenon of over-exploration. Our proposed method aims to improve optimization performance in high-dimensional spaces by adapting to the search domain and specifically discretizing it in promising regions. To achieve this, we utilize Markov Chain Monte Carlo (MCMC) for BO, which is a widely-used technique known for its ability to effectively sample from high-dimensional posterior distributions. We introduce \algo, a method that transitions candidate points from their original positions towards the approximated stationary distribution of TS. By only tracking a batch size number of points during transitions, \algo~significantly reduces the storage requirements in comparison to an entire mesh while still maintaining theoretical performance guarantees. It can be easily generalized to different scenarios with few additional hyper-parameters. Based on the basic MH version of \algo~. We also propose the Langevin dynamics (LD) version which transits following the gradient of the log-likelihood. Overall, \algo~can serve as a versatile algorithm that can be linked to most existing BO methods providing a posterior for any candidate points for continuous problems. We summarize our contributions as follows:
\begin{itemize}[leftmargin=*,nosep]
  \item We propose \algo, a Bayesian optimization algorithm which performs adaptive local optimization in high-dimensional problems that achieves both time and space efficiency.
  
  \item We provide theoretical guarantees on the convergence of \algo. To our knowledge, this is the first regret bound on high-dimensional Bayesian optimization problem which can deal with the scaling of dimensions with limited candidate points per round and avoid the overuse of memory.
  
  \item We experimentally show that \algo~, combined with both MH and LD, outperforms other strong baselines on various high-dimensional tasks. 
\end{itemize}
\vspace{-10pt}
\section{Related Work}
The optimization of black-box functions has been broadly used in many scenarios, such as hyper-parameter tuning and experimental design \citep{snoek2012practical, hernandez2017parallel}. Evolutionary algorithms are a class of methods with a long history and good performance to tackle the black-box optimization problem. One representative is CMA-ES, which adaptively adjusts the covariance matrix to generate new samples \citep{hansen2003reducing}.  These kinds of problems can also be formalized under the Bayesian optimization framework \citep{shahriari2015taking, frazier2018tutorial}, which is also more efficient. However, conventional BO algorithms are limited to low-dimensional cases and a small number of observations. There are two major components in BO: modeling and performing acquisition. 
To improve scalability and efficiency, various methods have been proposed for these two aspects. 
Regarding modeling, there exist a large amount of works which scale BO to higher dimensions via approximated GP\citep{hensman2013gaussian}, dimension reduction\citep{nayebi2019framework} or function decomposition \citep{wang2018batched}.
We leave the detailed discussion of this part in the appendix.


In the acquisition phase, balancing the exploration-exploitation trade-off is important. High-dimensional optimization problems often suffer from inadequate exploitation. To prevent over-exploring uncertain points near the domain boundary, methods like strong priors, dimension dropout, and cylindrical kernels have been proposed to allocate more sample points in the center region \citep{eriksson2021high, li2018high, oh2018bock}. Full exploration in high-dimensional space requires unrealistic computational resources and time.  Therefore, adequate exploitation of optimal points determines final performance. Previous algorithms for optimizing high-dimensional spaces include TuRBO and LA-MCTS. TuRBO limits the candidate set to a small box around the best points and adjusts the box based on later evaluations \citep{eriksson2019scalable}.LA-MCTS partitions the region with a constructed tree, recommending promising regions to improve algorithm performance \citep{wang2020learning}. Both methods focus on dividing promising regions, but maintaining a high-precision discretized grid for such regions is costly and can limit computation storage. The trade-off between regret and storage emphasizes the need for improvements in current methods. Therefore, we propose attention to the ability to congregate candidate points around the optimum in our algorithm.

Focusing on probabilistic acquisition, which involves using stochastic samples from the model posterior, sampling from the posterior distribution can be a bottleneck, especially for problems without a closed form expression for its posterior distribution. To address this, approximate sampling techniques are commonly used to generate samples from posterior approximations \citep{chapelle2011, Vanroy2017}. These approaches have shown effective practical performance \citep{Riquelme2018DeepBB,Urteaga2017VariationalIF}. However, maintaining performance over arbitrary time horizons while using approximate sampling remains unclear. Recent works have explored the application of LD, a stochastic sampling approach, in Multi-armed Bandit (MAB) problems \citep{mazumdar20a, Xu2022Langevin}. Instead of restricting to a parameterized utility function and optimizing the parameter space, we directly employ LD on the continuous and infinite action space. We believe this approach is more general and effective as it allows direct acquisition of candidate points without the need for optimization in the high-dimensional space.

\section{Background}

\subsection{Modeling with Gaussian processes}
\label{sec: GP intro}
Black-box optimization under uncertainty aims to find 
\begin{equation}
\small
\mathbf{x}^{*} \in D \textrm{ such that } f\left(\mathbf{x}^{*}\right) \geq f(\mathbf{x}), \quad \forall \mathbf{x} \in D
\end{equation}
where $f: D \rightarrow \mathbb{R}$ is an unknown function. In BO, the unknown objective $f$ is viewed as a probability distribution, and GP emerges as an effective way to estimate the unknown function, maintain an uncertainty estimate, and can be sequentially updated with available information. A GP is characterized by its mean function $\mu(\cdot)$ and kernel $k(\cdot, \cdot)$. The prior distribution of $f(x)$ is assumed to be $\mathcal{N}(0, k(x, x))$. Under this prior, the sample points $A_T := [x_1, \cdots, x_T]$ and the observations $[y_1, \cdots y_T]$ follows the multivariate distribution $\mathcal{N}(0, K_T + \sigma^2 I)$, where $K_T = [k(x, x')]_{x, x' \in A_T}$. Denote $k_T(x) = [k(x_i, x)]^T_{i=1:T} \in \mathbb{R}^{T \times 1}$, the posterior distribution over $f$ is thus Gaussian with mean $\mu_T(x)$ and covariance $k_T(x, x')$ that satisfy:
$ \mu_T(x)= k_T(x)^T(K_T + \sigma^2 I)^{-1}y_T,   k_T(x, x') = k(x, x') - k_T(x)^T (K_T + \sigma^2 I)^{-1}k_T(x')
$
\vspace{-10pt}
\subsection{Gaussian process and Thompson sampling}
In online learning tasks where the actions $x_t$ have to be sequentially chosen at each time step $t$, a common metric for measuring the performance is regret, which is defined using the optimal action $x^* := \text{arg max}_{x \in D} f(x)$ such that $r_t := f(x^*) - f(x_t)$ for the instantaneous regret at time $t$. The goal of optimization is to minimize the sum of  $r_t$, which entails finding optimal points while effectively exploring the sample space. To achieve this, we build upon ideas inspired by TS, which is to select the best point using a stochastic sample from the most recent model. TS is a classical MAB algorithm that can balance the exploitation-exploration trade-off. In every round, the algorithm estimates the reward of each arm by sampling from their respective distributions and plays the arm with the highest estimate. Thus, exploitation of optimal arms is achieved by heuristically playing the optimal arm, while exploration is guaranteed by the intrinsic uncertainty of probabilistic sampling. With an evolving GP, the algorithm first samples $f_t \sim GP(\mu_t(x), k_t(x, x'))$ and selects $x_t := \text{argmax}_{x \in \Omega} f_t(x)$ as the candidate points. Chowdhury and Gopalan gave a theoretical analysis for the algorithm \textsc{GP-TS}, which combines GP and TS to optimize stochastic MAB problems \citep{chowdhury2017kernelized}.
\vspace{-10pt}
\subsection{Markov Chain Monte Carlo}
\label{sec: MCMC intro}
Markov Chain Monte Carlo is used in GP modeling to approximate intractable probability distributions \citep{hensman2015mcmc}. Similarly, we hope to utilize MCMC to sample from a distribution on the action set corresponding to our acquisition approach. \par

\textit{Metropolis-Hastings} (MH) is an MCMC sampling algorithm that can sample from a target distribution $\pi(x), \; x \in \mathcal{X}$ known up to a constant, i.e. when we have knowledge of $\pi_d(x) = c \cdot \pi(x)$ \citep{metropolis1953equation}. 
With a given proposal distribution $q(\cdot , x)$, the algorithm samples a candidate point $y \in \mathcal{X}$ given the current value $x$ according to $q(y ,x)$. Then the Markov Chain transitions to $y$ with acceptance probability $\alpha(x, y) = \min \left\{1,  \frac{\pi_d(y)q(y, x)}{\pi_d(x)q(x,y)} \right\}$. This ratio yields a high acceptance probability when the proposed point is likely to be a better choice than the current point, so the points transit towards regions with higher function values. The MH algorithm is guaranteed to converge to a stationary distribution that is exactly $\pi(\cdot)$. 

\textit{Langevin dynamics} (LD), known to converge to the steady state distribution $\pi(x)$, provides another efficient iterative sampling procedure. From any arbitrary point $x_0$ sampled from a prior distribution, the update rule is given by $x_{i+1} = x_{i} + \epsilon \cdot \nabla \log{\pi(x)} + \sqrt{2\epsilon} z_i, \quad z_i \sim \mathcal{N}(0, I).$

Directly applying these two MCMC algorithms is challenging, as we do not have access to the complicated target distribution $\pi(x)$ induced by our probabilistic acquisition approach. In Section~\ref{sec: algorithm design}, we propose estimates for the acceptance probability $\alpha(x, y)$ and $\nabla \log{\pi(x)}$ that allows us to sample from the posterior distribution.



\section{Algorithm Design}
\label{sec: algorithm design}
We propose \algo, a novel algorithmic framework that combines MCMC methods with GP to evaluate more points effectively while maintaining a dense, time-varying discretization near promising regions for optimization.

The key idea of our algorithm is inspired by TS in which points with larger probability of being optimal are chosen next.  However, probabilistic acquisition in this manner is computationally demanding due to the required size of the discretization set and the complexity of the induced distribution.  Coarse partitioning of the input domain naturally leads to regret, as the best possible point in the mesh may still be far from the optimum. Consequently, a large number of points is necessary to ensure a fine partition. In high-dimensional spaces, this need for numerous points can result in large matrices for sampling in stochastic acquisition approaches and slow predictions.This presents a dilemma: a coarse mesh leads to significant regret, while a fine mesh makes sampling unrealistic. To efficiently sample from an intractable distribution and identify better candidate points, we employ MCMC. \par 

\algo~offers great flexibility in choosing different MCMC algorithms to sample from the GP posterior. We provide two MCMC subroutines using MH and LD. Our proposed method is compatible with any existing BO algorithms with probabilistic acquisition functions over a continuous domain (see Appendix).  The MH algorithm introduced in Section~\ref{sec: MCMC intro} can serve as a overarching framework using various acceptance ratios, and we propose an acceptance ratio that approximates the transition probability induced by the acquisition approach.
            

            

\begin{algorithm2e}
    \DontPrintSemicolon
    \LinesNumbered
    \SetAlgoLined
    \caption{\algo}
    \label{alg: MCMC-BO}
    \KwIn{Initial dataset $\mathcal{D}_0$, batch size $m$, MCMC transition number $N$, MCMC transition parameter $\Theta$}
    \For{$t = 0, 1, \cdots$}{
        Update posterior distribution GP$\big(\mu_t(x), k_t(x, x')\big)$ on $f$ using $\mathcal{D}_t$\;
        Create discretized candidate points $\boldsymbol{x}_t^0$ from continuous search domain\;

        \tcp{MCMC transition start}
        \For{$i = 0$ \KwTo $N-1$}{
            \For{$k = 0$ \KwTo $m-1$}{
                Perform \textbf{[MCMC routine]} using GP$\big(\mu_t(x), k_t(x, x')\big)$ on $x_{tk}^i$ with parameter $\Theta$\;
            }
        }

        \tcp{MCMC transition end}
        Observe $\textbf{\emph{y}}_t = f(\textbf{\emph{x}}_t^N)$\;
        $\mathcal{D}_{t+1} \leftarrow \mathcal{D}_t \cup (\textbf{\emph{x}}_t^N, \textbf{\emph{y}}_t)$\;
    }
\end{algorithm2e}

TS implicitly induces a distribution on each candidate point $x \in D_t$, with the probability of being selected as $\PP(f_t(x) \geq f_t(x')), \forall x' \in D_t$. Let $x_p, x_o$ be two candidate points after $n$ samples denoted by history $\mathcal{H} = \{(x_t, y_t)\}_{t = 1}^n$. We can sample according to the distribution of TS using MH with an acceptance rate $\alpha = \min \left\{1,  \frac{\PP(f_t(x_p) \geq f_t(x),\forall x \in D_t ) \cdot q(x_p \mid x_o)}{\PP(f_t(x_o) \geq f_t(x),\forall x \in D_t ) \cdot q(x_o \mid x_p)} \right\},$ where, $D_t$ represents the discretization set of the domain $D$ at time $t$, and $x_p$ is determined by the transition kernel $q(\cdot \mid \cdot)$. However, computing the probability involved in the ratio, $\PP\big(f_t(x_p) \geq f_t(x), \forall x \in D_t\big)$, requires marginalizing over the joint distribution of a high-dimensional multivariate normal distribution, which is intractable. Therefore, we propose an alternative approach that captures a similar idea. We define the acceptance probability as $\alpha = \min \left\{1,  \frac{\PP(f_t(x_p) \geq f_t(x_o)) q(x_p| x_o)}{\PP(f_t(x_o) \geq f_t(x_p))q(x_o|x_p)} \right\}.$ From GP regression, we can view the surrogate function values $y_{p} = f_t(x_{p}), \; y_{o} = f_t(x_{o}) $ associated to selected points $\boldsymbol{x} := (x_{p}, x_{o})$ as a Gaussian random vector with joint distribution as specified in Section~\ref{sec: GP intro}. Using linear transformations of a Gaussian random vector, let $c^T = \begin{bmatrix} 1 & -1 \end{bmatrix}$, $ K_n = [k(x_{1:n}, x_{1:n})] \in \mathbb{R}^{n \times n},$ and $k_n(\boldsymbol{x}) = [k(x_{1:n}, x_p), k(x_{1:n}, x_o)] \in \mathbb{R}^{n \times 2}$, we have that $y_{p} - y_{o} \sim \mathcal{N}\left(c^T\mu, c^T \Sigma c \right)$, where
$c^T \mu = \big(k(x_p, x_{1:n}) - k(x_o, x_{1:n})\big)^T (K_n + \sigma^2 I)^{-1}y_{1:n},c^T \Sigma c  = \big(k(x_p, x_p) + k(x_o, x_o) - 2k(x_p, x_o)\big) - \big(k_n(\boldsymbol{x})c\big)^T (K_n + \sigma^2 I)^{-1} \big(k_n(\boldsymbol{x}) c\big).
$
Since $c^T\mu, c^T \Sigma c\in \mathbb{R}$, we have that $P(y_{p} - y_{o} > 0 | \mathcal{H}) = \Phi\left( \frac{c^T\mu}{c^T \Sigma c}\right)$ (here $\Phi(\cdot)$ denotes the CDF of the standard normal). Therefore, the acceptance probability reduces to
\begin{equation}
\small
\label{eq: acceptance prob.}
    \alpha =  \min \left\{1,  \frac{\Phi\left( \frac{c^T\mu}{c^T \Sigma c}\right)  q(x_p| x_o)}{(1-\Phi\left( \frac{c^T\mu}{c^T \Sigma c}\right))q(x_o|x_p)} \right\}.
\end{equation}
\setlength{\intextsep}{0pt}
\setlength{\columnsep}{18pt}
\begin{wrapfigure}{r}{0.5\textwidth}
\begin{algorithm2e}[H]
        \DontPrintSemicolon
        \LinesNumbered
        \SetAlgoLined
    \label{alg: MCMC-MH}
    \caption{\textbf{[MCMC routine] with Metropolis-Hastings}}
    \KwIn{GP posterior GP$\big(\mu_t(x), k_t(x, x')\big)$, point $x_o$, proposal distribution $q(\cdot \mid x) := \mathcal{N}(0, \sigma)$ with parameter $\Theta=\{\sigma\}$.}
    Sample $u \sim \text{Unif}[0,1]$\;
    Sample $x_{p} \sim x_o + \mathcal{N}(0, \Theta)$\;
    \If{$u \ge \min \left\{1,  \frac{\PP\big(f_t(x_p) \geq f_t(x_o)\big) q(x_p \mid x_o)}{\PP\big(f_t(x_o) \geq f_t(x_p) \big)q(x_o \mid x_p)} \right\}$}{
        $x_{p} \leftarrow x_o$ // Reject the transition\;
    }
\end{algorithm2e}
\end{wrapfigure}

A demonstration of \algo~ with MH is shown in Algorithm~\ref{alg: MCMC-MH}. \algo~prepares a batch of $m$ candidate points $x_{tk}^{i}$ each round either from direct discretization or points to be executed from other algorithms, where $t$ stands for the round number, $k= 1,2,3,\cdots,m$, and  $i$ stands for transition times. Then with a proposed transition kernel, often defined using Brownian motion $x^{i+1}-x^{i} \sim \mathcal{N}(0, \sigma)$, we generate $m$ pairs of points $x_{tk}^{i}$, $x_{tk}^{i+1}$. We accept or decline the transition with the ratio in Equation~\ref{eq: acceptance prob.}. The random walk of Markov chain enables dense discretization of continuous space on more optimal regions, as illustrated in Fig \ref{fig:algo}. 



\setlength{\intextsep}{0pt}
\setlength{\columnsep}{18pt}
\begin{wrapfigure}{r}{0.5\textwidth}
\begin{algorithm2e}[H]
        \DontPrintSemicolon
        \LinesNumbered
        \SetAlgoLined
    \label{alg: MCMC-Langevin}
    \caption{\textbf{[MCMC routine] with Langevin dynamics}}
    \KwIn{GP posterior GP$\big(\mu_t(x), k_t(x, x')\big)$, point $x$, $\Theta=\{\epsilon\}$ as Langevin transition step.}
    Sample $z \sim \mathcal{N}(0, 1)$\;
    Estimate $\nabla \log p_t(x)$ using (\ref{eq: grad_log})\;
    $x \leftarrow x + \epsilon \cdot \nabla \log p_t(x) + \sqrt{2\epsilon} \cdot z$\;
\end{algorithm2e}
\end{wrapfigure}



The generality of our framework also extends to other sampling methods such as LD, where the samples transit following Equation~\ref{eq: langevin}. Previously, MH aims to sample from the intractable distribution corresponding to our acquisition approach by simplifying the ratio between two discretization points. Denoting the time-varying density $p_t(x) \propto \PP(f_t(x) \geq f_t(x'),\forall x' \in D_t )$, LD updates requires computing $\nabla \log p_t(x)$. Similar to MH's acceptance ratio, we propose a simple estimate using the derivative of the GP posterior’s mean and covariance.



In our proposed simplification for \algo, we utilize MH to calculate the ratio between the "winning probabilities" of two discretization points (the probability of being $\arg \max$). This calculation is done specifically with respect to the posterior distribution on those two points, rather than the entire domain. The term $\nabla \log p_t(x)$ easily incorporates this simplified ratio as:
$\frac{\partial \log p_t(x)}{\partial x_i} = \frac{\frac{\partial p_t(x)}{\partial x_i}}{p_t(x)} = \lim_{h \rightarrow 0} \frac{\frac{p_t(x + e_i \cdot h) - p_t(x)}{h}}{p_t(x)} = \lim_{h \rightarrow 0} \frac{1}{h} \big( \frac{p_t(x + e_i h)}{p_t(x)} - 1 \big)$Thus, the approximation of the log-likelihood in LD has the following equation:
\vspace{-5pt}
\begin{equation}
\label{eq: grad_log}
    \frac{\partial \log p_t(x)}{\partial x_i} \approx \lim_{h \rightarrow 0} \frac{1}{h} \big( \frac{\mathbb{P}\big(f_t(x + e_i h) > f_t(x)\big)}{\mathbb{P}\big(f_t(x) > f_t(x + e_i h)\big)} - 1 \big)  \approx \frac{1}{h} \big( \frac{\textbf{p}_i(x,h)}{1 - \textbf{p}_i(x,h)} - 1\big),
    \vspace{-5pt}
\end{equation}

where $\textbf{p}_i(x,h) =\mathbb{P}\big(f_t(x+e_i h ) > f_t(x)\big) = \Phi(\frac{c^T \mu}{\sqrt{c^T \Sigma c}}).$ As in MH, this involves calculating the cumulative distribution function (CDF) of a bivariate Gaussian, with the difference being that $x_p \leftarrow x + e_i h$ and $x_o \leftarrow x$. Using this numerical differentiation of the likelihood ratio, we obtain the gradient of $\log p_t(x)$ and the LD form of \algo.

\begin{figure}[H] 
        \centering
		\includegraphics[width=\linewidth]{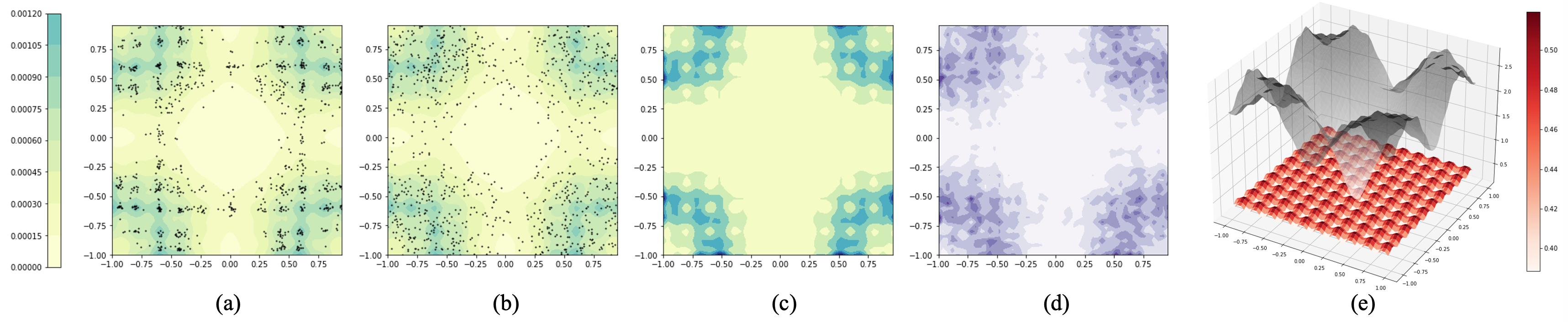}
  \vspace{-25pt}
    \caption{The figures are constructed from a $50 \times 50$ discretization of $D = [-1,1]^2$. \textbf{(a)(b)} The stationary distribution achieved by the MH and Langevin version of \algo, respectively, and the congregated points obtained after convergence of the transition process from current GP information. \textbf{(c)} TS distribution simulated using Monte Carlo. \textbf{(d)} standard deviation of TS distribution over 10 trials of $10^6$ samples. \textbf{(e)} GP posterior with surfaces being $\mu$ and $\Sigma$ on which \algo~transitions are performed.
    }
    \label{fig:stat}
\vspace{-10pt}
\end{figure}
\vspace{-10pt}
\section{Convergence Guarantee}
We provide convergence guarantees to our proposed algorithmic framework and derive the subsequent regret bounds. Our proposed algorithm removes the dependence on the size of the discretization set; whereas the original GP-TS setting requires that to achieve the theoretical upper bound on regret. As we only track a batch of $m$ points at a time, the algorithm complexity for \algo~only depends on $m$ and does not scale with the size of the discretization set $|D_t|$. 



\paragraph{Notation and assumption.} Let $\left\{x_{t}\right\}_{t=1}^{\infty}$ be an $\mathbb{R}^{d}$-valued discrete-time stochastic process adapted to
filtration $\left\{\mathcal{F}_{t}\right\}_{t=0}^{\infty}$. Denote $x^* = \text{argmax}_{x \in D} f(x)$, then the cumulative regret over the horizon $T$ is $R_T = \sum_{t=1}^T r_t$, where $r_t = f(x^*) - f(x_t)$. We also follow the convention and denote the maximum information gain at time $t$ as $\gamma_t$. Further notation is attached in Appendix B.
 
Leveraging the properties of Reproducing Kernel Hilbert Space(RKHS), we make a common assumption as in \citet{chowdhury2017kernelized}, that $f$ is Lipschitz constant with constant $B\cdot L$, where L is a constant associated with Kernel $k(\cdot,\cdot)$.\footnote{Further detail of assumption can be seen in Appendix B} In our analysis, we work with a discretization $D_t$ where $|D_t|$ is finite. We denote $[x]_{t} := \arg\min_{x' \in D_t} \| x - x'\|_2$. For convex and compact domain $D \subset [0, r]^d$, the size constraint $\left|D_{t}\right|=\left(B L r d t^{2}\right)^{d}$ guarantees that for every $x \in D$, $\left|f(x)-f\left([x]_{t}\right)\right| \leq\|f\|_{k} L\left\|x-[x]_{t}\right\|_{1} \leq 1 / t^{2}$. Note that the size of the discretization is only needed in the theoretical analysis and influences the regret bound, but the algorithm complexity is independent of it. Additionally, the transition kernel of MCMC is defined as $P(x, y) = Q(x,y)\alpha(x,y), \forall x,y \in D_t$, where $Q$ is the proposed distribution and $\alpha$ is the acceptance rate.

\subsection{Overview of Proof}
First, we show that the proposed approximated ratio of TS has a stationary distribution. Both MCMC algorithm (MH and LD) can converge to the stationary distribution with certain steps. Then the stationary distribution, though deviated from Thompson sampling still reserve the good properties. That is by defining a benign set, we can prove that the probability that the sampling points located in the benign set grows with $O(1-\frac{1}{T^2})$. An overview is presented here, and the detailed proof is deferred to Appendix C.
\begin{lemma}
\label{lem:lem1}
The proposed approximated posterior $P(x,\cdot)$, according to Alg.\ref{alg: MCMC-MH} and Equation.\ref{eq: acceptance prob.} does not yield a reversible Markov chain, but it still has a stationary distribution $\pi(x)$.
\end{lemma}
In Fig \ref{fig:stat} we demonstrate a 2D version of the stationary distribution of \algo~. Note that the acceptance probability our algorithm proposed is inspired by TS but does not lead to the same exact stationary distribution of TS. Furthermore, we show that the approximated probability $P(x, \cdot)$ converges to the stationary distribution.  In Sec~\ref{sec: exp} we also show that within a few hundred steps, the effect is significant enough. To obtain a bound on the overall regret, we begin by decomposing the instantaneous regret $r_t$ into two parts. From the choice of discretization sets $D_t$, $r_t$ depends on $\left[x^{\star}\right]_{t}$, which is the closest point to $x^\star$ in $D_t$. Therefore, we have that $r_{t} = \big(f\left(x^{\star}\right)-f\left(\left[x^{\star}\right]_{t}\right) \big) + \big(f\left(\left[x^{\star}\right]_{t}\right)-f\left(x_{t}\right) \big),$ where the difference $f\left(x^{\star}\right)-f\left(\left[x^{\star}\right]_{t}\right)$ is bounded by $1/t^2$ by the regularity assumption. We proceed to bound $f\left(\left[x^{\star}\right]_{t}\right)-f\left(x_{t}\right)$, which depends on the selected action $x_t$.

Regret increases when the approximation deviates from the underlying $f$. Thus, we hope that the function values of selected points $x_t$ and the approximation at each time step $t$ are not too far away. We define an event $E^f(t)$ and a benign set $G_t$ in Appendix C for the convenience of Lemma~\ref{lem: transition to bad}. The benign points set bound the difference between function value and at the $t-1$'s round GP's mean value.
\begin{lemma}
    \label{lem: transition to bad}
    For any filtration such that $E^f(t)$ is true, the transition probability from $[x^*]_t$ to any malignant point is bounded by 
    \begin{equation}
        P([x^*]_t,x') \leq exp(-(\frac{c_t}{2}-v_t)^2)(1-t^2), \quad \forall x' \in D_t \backslash G_t
    \end{equation}
\end{lemma}
Lemma~\ref{lem: transition to bad} shows that transiting on the stationary distribution $\pi(x)$, the probability of choosing a malignant points is small. 
With Lemma~\ref{lem: transition to bad}, we can further show that the probability of playing any action from $D_t \backslash G_t $ is small; the regret from undesirable action is then bounded with Theorem 5 in Appendix B.  
By giving the bound of the transition probability between the best point and the malignant points, the total regret can then be bounded in terms of standard deviation of the chosen actions $\sigma_{t-1}(x_t)$. With the information theoretic lemma (Lemma 3 in Appendix) we can then upper bound $\sum_{t=1}^T \sigma_{t-1}\left(x_t\right)$ with $ O\left(\sqrt{T \gamma_T}\right)$. As the difference $f\left(\left[x^{\star}\right]_{t}\right)-f\left(x_{t}\right)$ in the decomposition of $r_t$ is dependent on $\sigma_{t-1}(x_t)$, we finally arrive at the regret bound.

\begin{theorem}
Let $\delta \in(0,1), D \subset[0, r]^{d}$ be compact and convex, $\|f\|_{k} \leq B$,$\left\{\varepsilon_{t}\right\}_{t}$ a conditionally $R$-sub-Gaussian sequence. For any $T$, we let $N(T)$ to be the transition number for each round. Running \algo~ on a function $f$ lying in the RKHS $H_{k}(D)$ for $N(T)$ transitions per round and with decision sets $D_{t}$ chosen as above, with probability at least $1-\delta$, the regret of \algo~satisfies $R_T=O\left(\sqrt{\left(\gamma_T+\ln (2 / \delta)\right) d \ln (B d T)}\left(\sqrt{T \gamma_T}+B \sqrt{T \ln (2 / \delta)}\right)\right)$.
\end{theorem}
\section{Experiments}
\label{sec: exp}
In this section, we show the superiority of \algo~on both high-dimensional synthetic functions and Mujoco tasks. We also provide the ablation study for the algorithm in Table~\ref{tab:ablation-num} in Appendix.  We further validate that our proposed method itself is an effective GP-based BO algorithm as other MAB algorithms. We compare \algo~to state-of-the-art baselines of high-dimensional BO algorithms and EA algorithms.

For all BO algorithms, we utilize Thompson sampling to sample batches in each iteration and employ a scrambled Sobolev sequence to discretize the continuous search domain. The performance of \algo~is evaluated against TuRBO and LA-MCTS, which serve as its BO components. The performance figures illustrate the mean performance of the algorithms with one standard error.
\vspace{-10pt}
\begin{figure*}[htb] 
    \centering
    \includegraphics[width=1\linewidth]{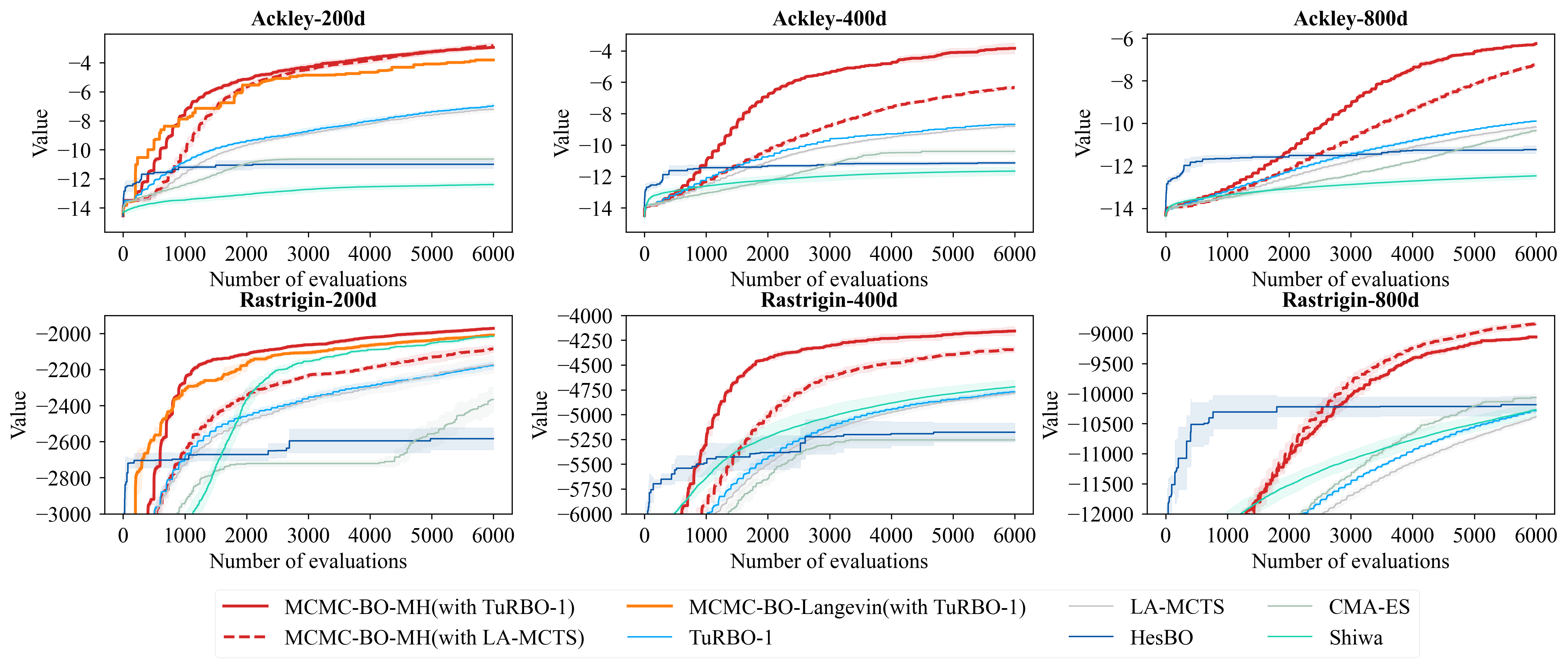}
    \vspace{-30pt}
    \caption{Optimization over high-dimensional synthetic functions.}
    \vspace{-10pt}
    \label{fig:syn}
\end{figure*}
\vspace{-10pt}
\subsection{High-Dimensional Synthetic Functions}
\begin{wrapfigure}{r}{0.5\textwidth}
 \label{fig:mujoco}
  \begin{center}
  \vspace{-10pt}
    \includegraphics[width=1\linewidth]{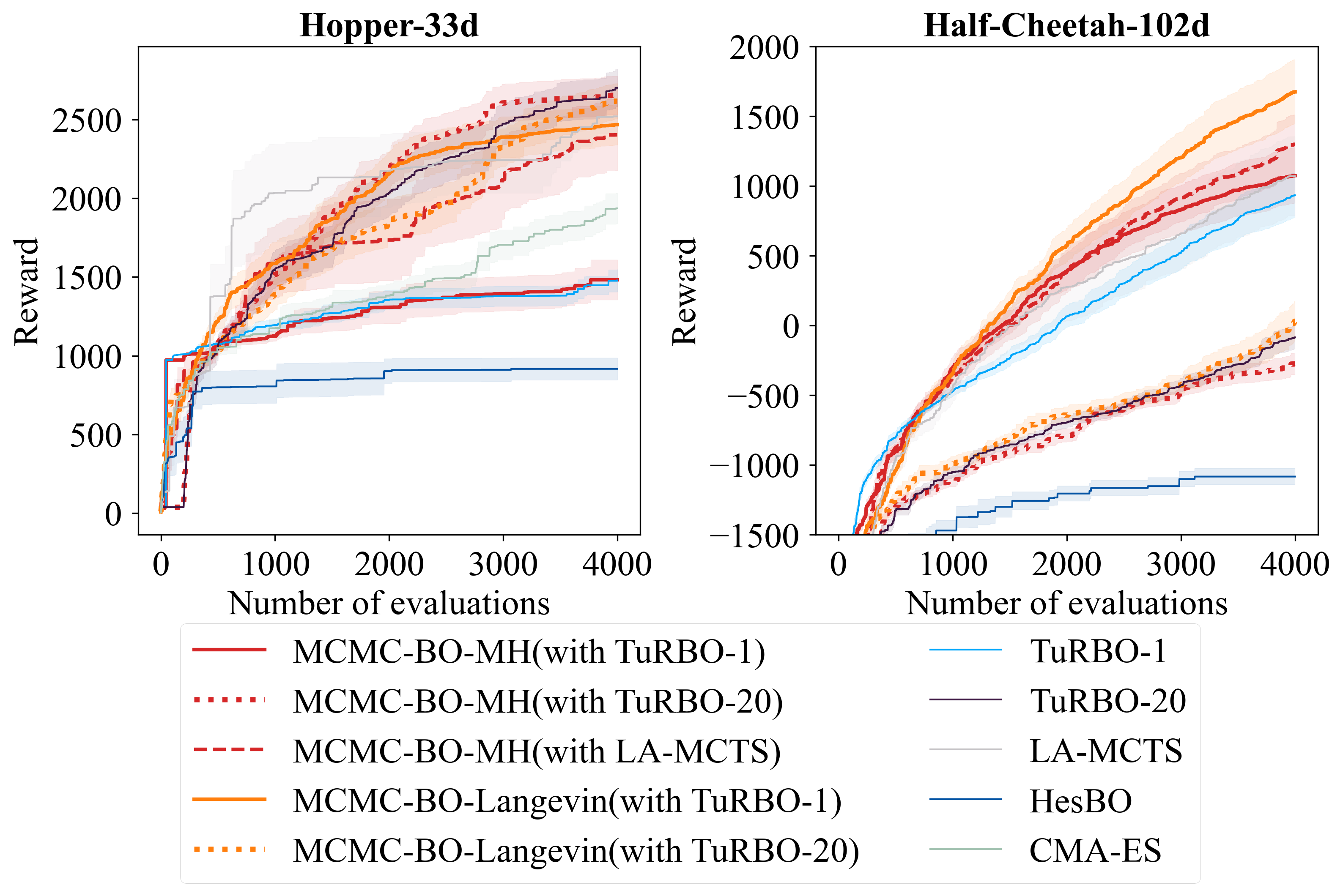}
    \vspace{-15pt}
  \end{center}
  \vspace{-20pt}
  \caption{Mujoco locomotion tasks.}
  \vspace{-20pt}
\end{wrapfigure}
\vspace{-10pt}
We selected two popular synthetic problems, Ackley and Rastrigin, to optimize. For the function dimension, we chose $d=200$, $400$, and $800$ to evaluate the performance in high-dimensional problems. We set the transition number to be the same as the function dimension. However, for the Langevin version of \algo, we ran only the $200d$ functions due to the lack of backward information for the posterior in the Gpytorch framework. All problems started with 200 initial points, and a batch of 100 points was sampled in each iteration.

Fig \ref{fig:syn} suggests that \algo~consistently outperforms other baselines on all functions. In higher dimensions $d = 800$, uniform discretization can not support good exploitation in such dimension, and TuRBO and LA-MCTS degenerate to the same level of performance as EA algorithms. \algo~achieves the best performance in all selected dimensions by allocating limited action points to more promising regions. 


\subsection{Mujoco Locomotion Task}
Mujoco locomotion tasks are widely used benchmarks for reinforcement learning algorithms \citep{todorov2012mujoco}. In our evaluation, we focus on Hopper and Half-Cheetah task, which have state spaces of dimensions 33 and 102, respectively. To assess the performance of the sampling-based algorithms, we optimize a linear policy: $\textbf{a} = \textbf{Ws}$ \citep{mania2018simple} , where the elements of the parameter matrix $\textbf{W}$ are continuous and range from $[-1, 1]$. The reward is computed over 10 episodes for each policy proposal. Both tasks start with 200 initial points and sample a batch of 50 points in each iteration. We set the transition number to 200 on both tasks. Fig \ref{fig:mujoco} shows the optimization performance of all algorithms. In the Hopper task, \algo with TuRBO-20 algorithm converges faster than the original TuRBO-20. 
In higher dimensional Half-Cheetah task, \algo~with TuRBO-1 still outperforms other baselines.  
\vspace{-10pt}



\subsection{Performance on low-dimensional problems}
\begin{wrapfigure}{r}{0.5\textwidth}
\vspace{-10pt}
  \begin{center}
    \includegraphics[width=1\linewidth]{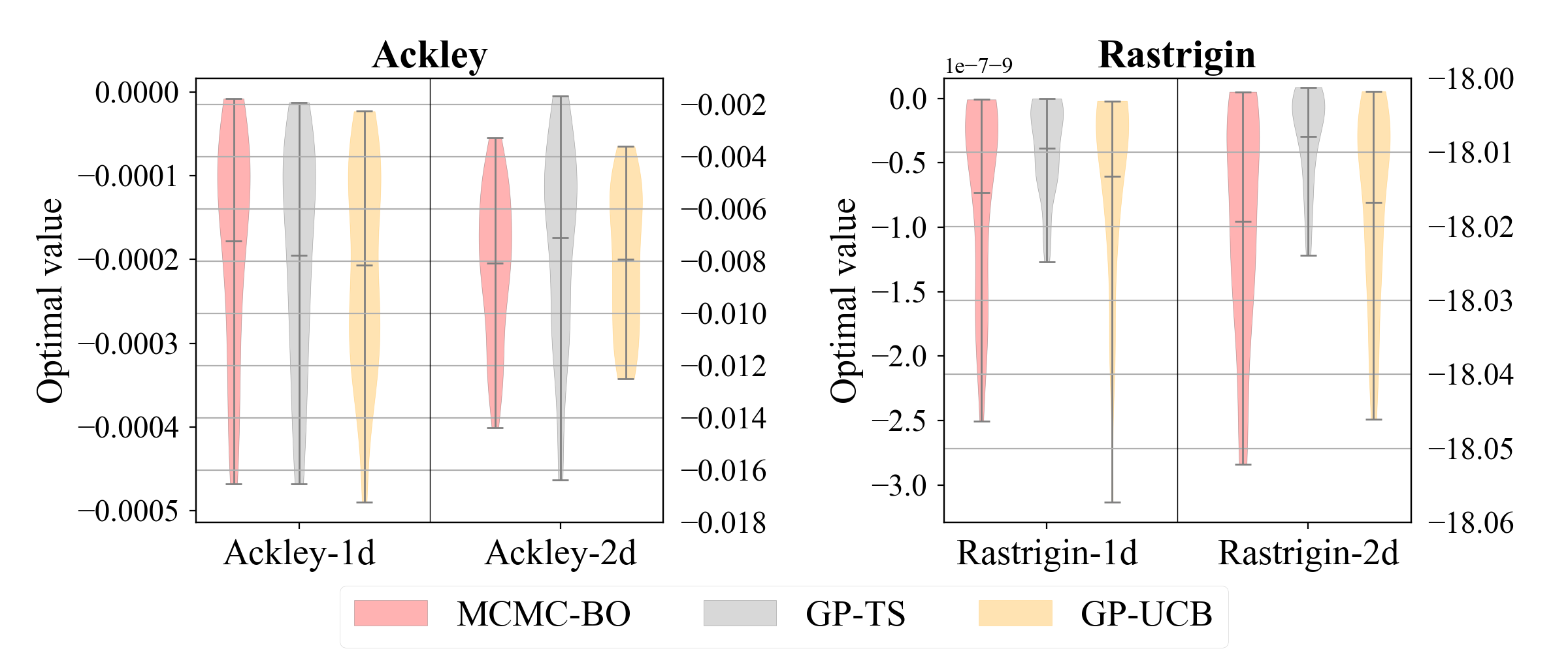}
  \end{center}
  \vspace{-20pt}
  \caption{We demonstrate the convergence of \algo~ compared with others on 1d and 2d functions.}
  \vspace{-10pt}
  \label{fig:best_ans}
\end{wrapfigure}
We compare \algo~with two popular BO algorithms, GP-TS and GP-UCB, on 1-2d synthetic functions. Figure \ref{fig:best_ans} numerically demonstrates the convergence of \algo. We depict the distribution over the optimal value encoded as a violin plot, with horizontal bars 20\% quantiles. GP-UCB only utilizes the diagonal information of the covariance matrix of $f(x_i)$, whereas TS uses the complete matrix information at once. As a compromise, \algo~employs a divide-and-conquer strategy by using a $2 \times 2$ sub-matrix at a time and still achieving an approximate stationary distribution of TS.


\section{Conclusion and Future Work}
Sequential optimization in high-dimensional spaces has a profound impact on machine learning. In this paper, we propose \algo~as a solution to improve the sample efficiency of high-dimensional BO. Our algorithm \algo~offers versatile transitions to promising regions instead of maintaining a huge candidate set. \algo~can be viewed as a GP-based bandit algorithm that yields an effective approximation for the induced TS distribution without the need to invert matrices of thousands of dimensions. We derive the regret bound of \algo~under high-dimensional cases without memory overuse. We also conduct comprehensive evaluations to show that \algo~can improve on existing popular high-dimensional BO baselines. 

In future research, we aim to develop parallel computation mechanisms to further enhance computational speed. Additionally, implementing analytical backward computations holds the potential for significant acceleration. It is worth noting that \algo~ can transit in different irregular spaces. We also look forward to combining \algo~with more complex space partition algorithms.



\bibliography{citations}

\end{document}